\documentclass{article}
\usepackage[dblblindworkshop, final]{neurips_2026}
\workshoptitle{Sim2Science}
\usepackage[protrusion=false,expansion=false]{microtype}
\usepackage[utf8]{inputenc}
\usepackage{float}
\usepackage{hyperref}
\usepackage{url}
\usepackage{booktabs}
\usepackage{array}
\usepackage{amsmath}
\usepackage{amssymb}
\usepackage{amsfonts}
\usepackage{xcolor}
\usepackage{graphicx}
\usepackage{tikz}
\usepackage{algorithm}
\usepackage{algorithmic}
\usetikzlibrary{arrows.meta,positioning,fit,backgrounds}

\title{Low-Fidelity FDM Spectral Guidance for Neural Eigenvalue Solvers}

\author{
\small
Aryan Chaudhary \\ BITS Pilani \\ \texttt{\scriptsize f20230651@pilani.bits-pilani.ac.in}
\And
Manikandan Padmanaban \\ IBM Research \\ \texttt{\scriptsize manipadm@in.ibm.com}
\And
Jagabondhu Hazra \\ IBM Research \\ \texttt{\scriptsize jahazra1@in.ibm.com}
}

\newcommand{\Lop}{\mathcal{L}}
\newcommand{\Id}{I}
\newcommand{\inner}[2]{\left\langle #1,#2\right\rangle}
\newcommand{\norm}[1]{\left\lVert #1\right\rVert}

\begin{document}

\maketitle

\begin{abstract}
Operator eigenvalue problems appear throughout science. Classical methods
usually discretize the operator into a matrix and then solve the resulting
matrix eigenvalue problem. This works well in low dimensions, but fine grids
quickly become expensive in both memory and computation as the dimension
grows. Neural network based solvers avoid storing these large grids, but recent
state of the art neural methods can require hundreds of thousands of training
steps and may struggle to find the desired eigenvalues. We show that the two
approaches can help each other. A coarse finite difference method (FDM)
calculation acts as a cheap numerical model of the operator spectrum. We use the approximate eigenvalues as fixed shifts during the training of the neural solver, as
  they only need to locate the relevant part of the spectrum.
We also introduce
Stabilized Inverse Power Method Neural Network (SIPMNN), a more stable training
procedure for higher-dimensional problems. Across five test problems at
$d=10$, the combined approach is more accurate overall than the tested fully neural
  alternatives while using eight to ten times fewer iterations.
  
\end{abstract}

\section{Introduction}

We study the low spectrum of a differential operator,
\begin{equation}
    \Lop u_i=\lambda_i u_i, \qquad i=1,\ldots,k.
\end{equation}
Finite difference and finite element methods turn the continuous operator into
a matrix eigenproblem. They work well in low dimensions, but $n$ points in
each of $d$ coordinates give $n^d$ grid points
\citep{leveque2007finite}. Neural eigensolvers instead
represent $u_i$ with a neural network
and apply $\Lop$ with automatic differentiation
\citep{yu2018deepritz,yang2023pmnn}. They avoid storing a fine tensor grid, but
they replace a matrix eigensolve with a difficult optimization problem. Without
an external estimate of the relevant spectral region, the neural method must
also discover that information during training.
This suggests a natural hybrid setting. A lower-fidelity numerical
model may be too inaccurate to serve as the final solver, yet still provide
useful structural information about where the desired eigenvalues lie.

Inverse power methods depend strongly on the shift $\sigma$. They amplify each
eigencomponent by $(\lambda_i-\sigma)^{-1}$. Inverse Power Method Neural Network (IPMNN) showed that a supplied shift
near a desired eigenvalue can be used to find that eigenpair
\citep{yang2023pmnn}. However, the shift was treated as prior knowledge. The
paper did not give a general, practical way to obtain it. STNet instead updates
the shift during training and combines it with a filter containing several factors
\citep{wang2025stnet}. This removes the need for an external numerical shift,
but its reported experiments use 400,000 iterations at $d=1,2$ and 500,000
iterations at $d=5$.

Our first question is whether a coarse discretization can
provide useful spectral guidance even when its eigenvalues are not accurate
enough to serve as final solutions.
We solve the coarse matrix eigenproblem once and use
its rough eigenvalues as fixed shifts. SIPMNN then refines the eigenpairs
without a fine tensor grid.
We keep the learned deflation from STNet, but drop
its adaptive filter with several factors. Each mode instead uses the simpler
IPMNN filter with one fixed FDM value. Our second question concerns stability
in higher dimensions. Basic IPMNN training can remain unstable even with a
useful shift, which motivates the stabilized method SIPMNN.

Our contributions are twofold. First, we show how a coarse numerical solve can
guide a neural eigensolver.
Its estimates provide useful fixed values of
$\sigma$ and reduce the training budget from the
400,000 to 500,000 iterations reported for STNet to 50,000 iterations on
several problems.
Second, we introduce three changes to IPMNN, namely a smooth
initial target, delayed target updates, and a bounded normalized penalty.
We test their roles with a matched ablation at $d=10$. In the controlled
$d=10$ comparison, Fixed FDM SIPMNN gives lower or tied eigenvalue error than
both fully neural configurations with adaptive shifts for all ten reported modes.
Together, these results support a hybrid strategy in which a
coarse numerical model supplies spectral guidance and SIPMNN solves the
higher-dimensional eigenproblem. Under the matched 50,000-step comparison,
this combination gives lower or tied error than the tested fully neural
configurations for all ten $d=10$ modes.

\section{Background and Related Work}

Classical eigensolvers use Krylov subspaces, Rayleigh--Ritz projection,
deflation, filters, and shifted inverse methods
\citep{golub2013matrix,saad2011eigenvalue,banerjee2016chebyshev}. If
$\Lop v_i=\lambda_i v_i$, then
$(\Lop-\sigma\Id)^{-1}v_i=(\lambda_i-\sigma)^{-1}v_i$, so the eigenvalue
closest to $\sigma$ gets the largest response.

Neural PDE solvers represent trial functions with neural networks. Deep Ritz
methods minimize an energy \citep{yu2018deepritz}. PMNN and IPMNN turn power
and inverse power iteration into neural methods \citep{yang2023pmnn}. STNet
combines inverse iteration, learned deflation, adaptive shifts, and a
filter with several factors \citep{wang2025stnet}. Its reported experiments use
400,000 iterations at $d=1,2$ and 500,000 iterations at $d=5$. Other neural
spectral methods learn several modes or use low rank objectives
\citep{pfau2019spin,deng2022neuralef,ryu2024operatorsvd,benshaul2023eigenvalue}.
Here we retain the inverse iteration setting and ask whether a coarse FDM solve
can provide a useful fixed shift.

\begin{figure}[t]
\centering
\begin{tikzpicture}[
    font=\small,
    node distance=2.5mm,
    flowbox/.style={draw=blue!55!black, fill=white, rounded corners=3pt,
                 align=center, minimum height=19mm, text width=18mm,
                 line width=.65pt, inner sep=1.2mm},
    prior/.style={flowbox, draw=teal!65!black, fill=teal!8, text width=42mm},
    solver/.style={draw=orange!70!black, fill=orange!10, rounded corners=4pt,
                   align=center, minimum height=19mm, text width=40mm,
                   line width=.8pt, inner sep=1.2mm},
    output/.style={draw=violet!65!black, fill=violet!9, rounded corners=4pt,
                   align=center, minimum height=19mm, text width=20mm,
                   line width=.8pt, inner sep=1.2mm},
    arrow/.style={-{Latex[length=2.2mm]}, line width=.75pt,
                   draw=gray!75!black},
    feedback/.style={-{Latex[length=2.2mm]}, dashed, line width=.7pt,
                      draw=violet!65!black}
]
\node[flowbox] (operator) {Operator $\Lop$\\and domain};
\node[prior, right=of operator] (prior)
    {\textbf{Coarse prior, run once}\\
     FDM with $n=5$\\    
    {$\widehat\lambda_i^{\mathrm{FDM}}\ \longrightarrow\
      \sigma_i=\widehat\lambda_i^{\mathrm{FDM}}-1$}};
\node[solver, right=of prior] (sipmnn)
    {\textbf{SIPMNN for mode $i$}\\
     fixed-shift inverse loss\\
     smooth target $\boldsymbol{\cdot}$ delayed updates\\
     bounded, normalized penalty};
\node[output, right=of sipmnn] (pairs)
    {\textbf{Refined mode}\\
     $(\widetilde\lambda_i,\widetilde u_i)$};

\draw[arrow] (operator) to (prior);
\draw[arrow] (prior) to (sipmnn);
\draw[arrow] (sipmnn) to (pairs);
\draw[feedback] (pairs.south) -- ++(0,-5mm) -|
    node[pos=.55, below, font=\scriptsize, text=violet!65!black]
    {deflation for higher modes} (sipmnn.south);
\end{tikzpicture}
\caption{SIPMNN first obtains rough eigenvalues from one coarse FDM solve.
Their shifts stay fixed while the neural networks refine each mode. Learned
deflation passes each lower-mode estimate to the higher-mode updates.
}
\label{fig:pipeline}
\end{figure}
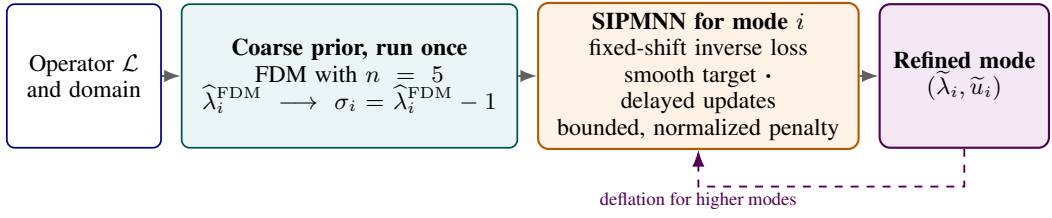

\section{Fixed Shifts and SIPMNN}

\subsection{Neural inverse iteration}

Let $u_\theta$ be a scalar neural network evaluated at $N$ collocation points.
For a target function $u_{\mathrm{old}}$, we minimize
\begin{equation}
\mathcal L_{\mathrm{IPM}}=
\left\|
\frac{(\Lop-\sigma\Id)u_\theta}
{\norm{(\Lop-\sigma\Id)u_\theta}}-u_{\mathrm{old}}
\right\|_2^2.
\label{eq:ipm-sim}
\end{equation}
The eigenvalue is estimated independently with the Rayleigh quotient
\begin{equation}
\lambda_{\mathrm{RQ}}(u_\theta)=
\frac{\inner{u_\theta}{\Lop u_\theta}}
{\inner{u_\theta}{u_\theta}}.
\label{eq:rq-sim}
\end{equation}
We measure whether the output satisfies the eigenvalue equation using the
normalized residual
\begin{equation}
r(u_\theta)=
\frac{\norm{\Lop u_\theta-\lambda_{\mathrm{RQ}}u_\theta}_2}
{\norm{u_\theta}_2}.
\label{eq:residual-sim}
\end{equation}
This is a diagnostic rather than a training loss. A small residual shows that
the network found an eigenpair, while the reference eigenvalue shows whether
it found the intended one.
This distinction matters because a coarse shift may lead the
neural solver to a valid eigenpair other than the one intended.
\subsection{Fixed coarse FDM shifts}
For the runs with fixed shifts, we first compute rough eigenvalues
$\widehat\lambda_i^{\mathrm{FDM}}$ with a coarse FDM grid of five interior
points per coordinate axis, and set
\begin{equation}
    \sigma_i=\widehat\lambda_i^{\mathrm{FDM}}+\delta,
    \qquad \delta=-1.
\end{equation}
The coarse solve is run once and $\sigma_i$ remains unchanged during neural
training.

\subsection{Smooth target initialization}

The usual IPMNN and STNet initialization assigns independent random values to
the collocation points. Nearby points can therefore have unrelated targets.
Because the network represents a continuous function, early training is spent
fitting this rough pointwise noise. SIPMNN instead starts from a random smooth
function containing only low frequencies. This target is easier to represent,
but contains no reference eigenfunction or shape chosen for the operator. The
shift and inverse iteration still determine which component is amplified.



\subsection{Delayed target updates}
IPMNN and STNet replace $u_{\mathrm{old}}$ with the normalized network output
after every optimizer step. Since $u_{\mathrm{old}}$ depends on the network's
own output, the target shifts as soon as the network updates, so training can
chase a moving target rather than converge toward a fixed shifted-inverse
step. SIPMNN instead holds $u_{\mathrm{old}}$ fixed for 500 steps, giving Adam
time to make progress on a stable objective before the target is refreshed.
\subsection{Bounded normalized penalty}

We guide the Rayleigh quotient toward the coarse estimate with
\begin{equation}
\mathcal L_\sigma^{\mathrm{raw}}=w_\sigma
\left(\frac{\lambda_{\mathrm{RQ}}-\widehat\lambda_i^{\mathrm{FDM}}}{s_i}\right)^2,
\quad
s_i=\max(|\sigma_i|,|\widehat\lambda_i^{\mathrm{FDM}}|,1).
\end{equation}
We use $w_\sigma=10^{-6}$ and cap its value at
$5\mathcal L_{\mathrm{IPM}}$, giving
$\mathcal L=\mathcal L_{\mathrm{IPM}}+\mathcal L_\sigma$. The detached cap
is defined in Appendix~\ref{app:penalty-sim}. The penalty discourages movement
to a distant mode. Without it, the normalized IPM loss can be small at a
higher valid eigenpair even when that eigenvalue is far from the intended one.

An inaccurate coarse estimate can still point toward the wrong
spectral region, so the cap limits the penalty's influence without removing
this risk.

\subsection{Deflation and higher modes}

At each optimization iteration, we normalize the current estimates of the
lower modes and deflate mode $i$ using
\begin{equation}
D(\Lop)u=\Lop u-\sum_{j<i}\widetilde\lambda_j\widetilde u_j
\frac{\inner{\widetilde u_j}{u}}
{\inner{\widetilde u_j}{\widetilde u_j}}.
\label{eq:deflation-sim}
\end{equation}
All mode networks advance within the same training loop, and mode $i$ uses the
current estimates of modes $j<i$. Appendix~\ref{app:deflation-sim} states the
discrete normalization precisely. Errors in either the coarse shift or a
lower-mode estimate can therefore propagate through deflation to higher modes.
\begin{algorithm}[t]
\caption{Fixed FDM SIPMNN}
\label{alg:sipmnn-sim}
\begin{algorithmic}[1]
\setlength{\itemsep}{1pt}
\STATE \textbf{Input:} Operator $\Lop$, number of modes $k$, collocation set
$X$, iterations $K$, target-update interval $T_u$
\STATE \textbf{Output:} Eigenpairs
$(\widetilde\lambda_i,\widetilde u_i)$, $i=1,\ldots,k$
\STATE $\widehat\lambda_{1:k}^{\mathrm{FDM}}\leftarrow
\textsc{CoarseFDM}(\Lop,k,n=5)$
\STATE $\sigma_i\leftarrow\widehat\lambda_i^{\mathrm{FDM}}-1$ for all $i$
\FOR{$i=1$ to $k$}
    \STATE Initialize $u_{\theta_i}$, normalized smooth target
    $u_{\mathrm{old},i}$, and best loss $\ell_i\leftarrow\infty$
\ENDFOR
\FOR{$t=1$ to $K$}
    \STATE $\mathcal S_t\leftarrow\varnothing$
    \FOR{$i=1$ to $k$}
        \STATE Evaluate $u_{\theta_i}(X)$ and
        $(D_{\mathcal S_t}(\Lop)-\sigma_i\Id)u_{\theta_i}(X)$
        \STATE Compute $\lambda_{\mathrm{RQ},i}^{(t)}$,
        $\mathcal L_{\mathrm{IPM},i}^{(t)}$, and bounded
        $\mathcal L_{\sigma,i}^{(t)}$
        \STATE $\mathcal L_i^{(t)}\leftarrow
        \mathcal L_{\mathrm{IPM},i}^{(t)}+\mathcal L_{\sigma,i}^{(t)}$
        \STATE Update $\theta_i$ with Adam on
        $\mathcal L_i^{(t)}$
        \IF{$t\bmod T_u=0$}
            \STATE Replace $u_{\mathrm{old},i}$ with the normalized network
            output
        \ENDIF
        \IF{$\mathcal L_i^{(t)}<\ell_i$}
            \STATE Set $\ell_i\leftarrow\mathcal L_i^{(t)}$ and retain the current
            eigenpair as $(\widetilde\lambda_i,\widetilde u_i)$
        \ENDIF
        \STATE Append the current normalized eigenpair to $\mathcal S_t$
        for higher-mode deflation
    \ENDFOR
\ENDFOR
\STATE \textbf{Return:}
$(\widetilde\lambda_i,\widetilde u_i)_{i=1}^k$
\end{algorithmic}
\end{algorithm}

\section{Experimental Setup}

We use $\Lop=-\tfrac12\Delta+V(x)$ with Dirichlet boundary conditions and the
five benchmarks listed in Appendix Table~\ref{tab:benchmarks-sim}. We train
the first two modes at $d\in\{5,10\}$. Exact references are used when
available. Otherwise, we use an independent fine solve in one dimension or a
sine basis calculation. These references are separate from the coarse $n=5$
FDM values used for shifts.

We compare standard and stabilized training with either adaptive or fixed FDM
shifts.

All runs use fully connected neural networks with four hidden layers and tanh
activations, 50,000 static Latin hypercube
points, and 50,000 Adam steps at learning rate $10^{-3}$. The models for the
first two modes are updated within the same training loop. Standard IPMNN starts from pointwise
noise, updates its target after every step, and has no sigma penalty. SIPMNN
uses the three changes above. Table~\ref{tab:hyperparams-sim} gives the full
configuration.
The adaptive baseline is fully specified in Appendix~\ref{app:adaptive-sim}.

We report relative eigenvalue error against independent references.
Further discretization and evaluation details are in
Appendix~\ref{app:fdm-sim}.

\section{Results}

\begin{table*}[t]
\centering
\setlength{\tabcolsep}{1.5pt}
\renewcommand{\arraystretch}{1.16}
\caption{
Relative eigenvalue error (\%) after 50,000 neural optimization
steps, reported as $\lambda_1/\lambda_2$. The lowest error for each mode is
bold. The four configurations compare two training procedures and two choices
of spectral shift.}
\label{tab:results-sim}
\begin{tabular}{lcrrrr}
\toprule
& & \multicolumn{2}{c}{IPMNN} & \multicolumn{2}{c}{SIPMNN} \\
\cmidrule(lr){3-4}\cmidrule(lr){5-6}
Benchmark & Reference $\lambda_1/\lambda_2$ &
\shortstack{Adaptive\\shift} & \shortstack{Fixed FDM\\shift} &
\shortstack{Adaptive\\shift} & \shortstack{Fixed FDM\\shift} \\
\midrule
\multicolumn{6}{l}{$d=5$} \\
Anharmonic & 2.7957/4.0060 & \textbf{0.26}/1.17 & 0.43/4.02 & 2.20/12.21 & 0.78/\textbf{0.77} \\
Coupled oscillator & 2.4597/3.3541 & 3.70/2.92 & \textbf{3.12}/238.40 & 137.61/60.42 & 8.00/\textbf{1.10} \\
Double well & 4.3482/5.1403 & 0.10/\textbf{0.11} & 0.11/\textbf{0.11} & \textbf{0.05}/0.26 & 0.07/0.13 \\
Quantum harmonic & 2.5000/3.5000 & 3.14/5.63 & \textbf{2.45}/429.46 & 218.16/61.66 & 8.08/\textbf{1.80} \\
Random Gaussian & 24.6643/39.4399 & 0.01/37.46 & \textbf{0.00}/\textbf{0.00} & \textbf{0.00}/37.47 & \textbf{0.00}/0.09 \\
\midrule
\multicolumn{6}{l}{$d=10$} \\
Anharmonic & 5.5914/6.8018 & 666.70/132.52 & 197.31/198.36 & 346.86/176.89 & \textbf{1.38}/\textbf{1.81} \\
Coupled oscillator & 4.8616/5.7560 & 333.85/1922.16 & 143.81/146.61 & 279.32/190.92 & \textbf{28.06}/\textbf{5.10} \\
Double well & 8.6965/9.4885 & 166.83/85.67 & 76.51/90.93 & 187.46/152.43 & \textbf{0.53}/\textbf{1.50} \\
Quantum harmonic & 5.0000/6.0000 & 318.61/162.67 & 145.81/144.94 & 409.89/94.22 & \textbf{26.23}/\textbf{12.34} \\
Random Gaussian & 49.3480/64.1524 & 0.02/23.07 & \textbf{0.00}/23.09 & \textbf{0.00}/0.62 & \textbf{0.00}/\textbf{0.01} \\
\bottomrule
\end{tabular}
\end{table*}

Table~\ref{tab:results-sim} separates coarse guidance from stabilized training.
At $d=10$, fixed-FDM IPMNN and adaptive SIPMNN both have large errors, so
neither component alone is sufficient. Their combination works best, suggesting
complementary roles. FDM locates the useful spectral region, while SIPMNN
stabilizes the neural inverse iteration.

The $d=5$ problems are easier, and several configurations already work well,
so the advantage of the full combination is less decisive. Even there, fixed
FDM shifts improve or match adaptive shifts within SIPMNN in nine of ten modes.
The main benefit of combining both components becomes clearer at $d=10$, where
the simpler alternatives become unreliable.

The remaining failures follow the quality of the coarse estimate. The two
problems with the weakest first-mode results also have the least accurate coarse
first eigenvalues (Appendix Table~\ref{tab:quality-sim}). At the same time, the
strong improvements on the other problems show that SIPMNN does not simply
copy the FDM output. The coarse solve acts as a guide. When it locates the right
part of the spectrum, SIPMNN can refine it. When it points elsewhere, training
may settle on a valid but unintended mode.

The matched ablation in Appendix~\ref{app:ablation-sim} retains the same FDM
shifts and removes one SIPMNN change at a time. Removing the bounded penalty
raises mean error from 7.70\% to 77.33\%. Removing the smooth target raises it
to 8.85\%, and updating the target every step raises it to 11.46\%. The penalty
mainly prevents wrong-mode solutions, while the other changes give smaller but
useful improvements.

Appendix~\ref{app:residual-sim}
also reports normalized residuals as consistency diagnostics for the eigenpairs.

\section{Limitations}
We test at $d=5$ and $d=10$. Since its tensor grid scales as $n^d$, coarse FDM
becomes impractical at much higher dimensions even with five points per
coordinate. The method can also select the wrong mode, and first-mode errors
can propagate through deflation. A coarse estimate need not be highly accurate,
but it must identify the correct spectral region. Future work should study
shift estimators that avoid $n^d$ scaling and diagnostics that detect estimates
pointing to the wrong part of the spectrum.

\section{Conclusion}
A coarse FDM calculation provides rough spectral information from a
discretized matrix problem. It locates the relevant spectral region rather than
serving as the final solver, while SIPMNN uses its fixed shifts to refine the
eigenpairs without a fine tensor grid. This hybrid method gives the lowest or
tied error for every tested $d=10$ mode within 50,000 neural steps. The fully
neural adaptive variants do not match this accuracy, while STNet reports
400,000 to 500,000 steps for the corresponding experiments. The coarse values
need not be accurate final answers, but they must identify the useful spectral
region. Overall, the results support combining a cheap coarse numerical
calculation with a neural eigensolver.

\clearpage
\bibliographystyle{plainnat}
\bibliography{references}

\clearpage

\appendix
\section{Benchmark Definitions}
\label{app:benchmarks-sim}

\begin{table}[h]
\centering
\setlength{\tabcolsep}{5pt}
\renewcommand{\arraystretch}{1.08}
\caption{Schr\"odinger benchmark operators and domains.}
\label{tab:benchmarks-sim}
\begin{tabular}{lll}
\toprule
Benchmark & Domain & Potential $V(x)$ \\
\midrule
Anharmonic & $[-4,4]^d$ & $\frac12\sum_jx_j^2+0.1\sum_jx_j^4$ \\
Coupled oscillator & $[-5,5]^d$ & $\frac12[0.8\|x\|^2+0.2(\mathbf1^\top x)^2]$ \\
Double well & $[-2.5,2.5]^d$ & $\sum_j(x_j^2-1)^2$ \\
Quantum harmonic & $[-5,5]^d$ & $\frac12\|x\|^2$ \\
Random Gaussian & $[0,1]^d$ & Sum of eight Gaussian wells \\
\bottomrule
\end{tabular}
\end{table}

\section{Coarse-Shift Quality}

\begin{table}[h]
\centering
\setlength{\tabcolsep}{5pt}
\renewcommand{\arraystretch}{1.08}
\caption{Error of the coarse $n=5$ FDM estimate and the final SIPMNN result at
$d=10$, reported as $\lambda_1/\lambda_2$ in percent.}
\label{tab:quality-sim}
\begin{tabular}{lrr}
\toprule
Benchmark & Coarse FDM & Fixed FDM SIPMNN \\
\midrule
Anharmonic & 20.90/15.65 & 1.38/1.81 \\
Coupled oscillator & 35.38/21.89 & 28.06/5.10 \\
Double well & 10.87/12.01 & 0.53/1.50 \\
Quantum harmonic & 37.07/23.79 & 26.23/12.34 \\
Random Gaussian & 2.26/4.28 & 0.00/0.01 \\
\bottomrule
\end{tabular}
\end{table}

\section{Fixed FDM SIPMNN Ablation}
\label{app:ablation-sim}

All ablation runs use the following configuration:
\begin{itemize}
    \item \textbf{Dimension and modes:} $d=10$ with two modes.
    \item \textbf{Optimizer:} Adam with learning rate $10^{-3}$.
    \item \textbf{Neural network architecture:}
    $[d,80,80,80,80,1]$ with tanh activation.
    \item \textbf{Training points:} 50,000 static Latin hypercube points.
    \item \textbf{Iterations:} 50,000.
    \item \textbf{Spectral shifts:} the same coarse FDM shifts for every
    variant.
\end{itemize}

\emph{No smooth} uses independent pointwise noise, \emph{No delay} updates the
target after every optimizer step, and \emph{No penalty} sets $w_\sigma=0$.

\begin{table}[h]
\centering
\setlength{\tabcolsep}{3pt}
\renewcommand{\arraystretch}{1.15}
\caption{Fixed FDM SIPMNN ablation at $d=10$. Entries are relative errors
(\%) for $\lambda_1/\lambda_2$. The lowest error for each mode is bold.
}
\label{tab:ablation-sim}
\begin{tabular}{lrrrr}
\toprule
Benchmark & Full & \shortstack{No smooth\\target} &
\shortstack{No delayed\\update} & \shortstack{No sigma\\penalty} \\
\midrule
Anharmonic & \textbf{1.38}/\textbf{1.81} & 5.99/1.82 & 8.09/5.02 & 161.21/24.88 \\
Coupled oscillator & 28.06/\textbf{5.10} & 25.99/9.44 & \textbf{25.63}/10.61 & 124.92/5.14 \\
Double well & 0.53/1.50 & 1.21/\textbf{0.46} & \textbf{0.10}/1.46 & 66.71/98.45 \\
Quantum harmonic & \textbf{26.23}/12.34 & 29.67/13.93 & 30.05/\textbf{10.59} & 139.69/152.31 \\
Random Gaussian & \textbf{0.00}/\textbf{0.01} & \textbf{0.00}/\textbf{0.01} & \textbf{0.00}/23.07 & 0.01/0.02 \\
\bottomrule
\end{tabular}
\end{table}

The penalty has little effect on Random Gaussian but prevents large failures
on the other four operators. Across all ten modes, removing it raises mean
error from 7.70\% to 77.33\%. Excluding Random Gaussian, the mean over the
remaining eight modes rises from 9.62\% to 96.66\%. The variants without the
penalty completed the same 50,000-step budget. Their large errors are therefore
not an artifact of stopping early.

\section{Normalized Residuals}
\label{app:residual-sim}

\begin{table}[h]
\centering
\setlength{\tabcolsep}{12pt}
\renewcommand{\arraystretch}{1.08}
\caption{Normalized residuals $r_1/r_2$ for Fixed FDM SIPMNN at $d=10$.}
\begin{tabular}{lr}
\toprule
Benchmark & Residual $r_1/r_2$ \\
\midrule
Anharmonic & 0.078/0.124 \\
Coupled oscillator & 0.046/0.318 \\
Double well & 0.057/0.209 \\
Quantum harmonic & 0.064/0.057 \\
Random Gaussian & 0.032/0.395 \\
\bottomrule
\end{tabular}
\end{table}

These residuals check whether the output is close to an eigenpair. They do not
identify which eigenpair was found, so eigenvalue errors remain the main
metric for mode selection.

\section{Implementation Details}
\label{app:implementation-sim}
\subsection{Smooth initialization and delayed updates}

We rescale each coordinate to $[-1,1]$ and initialize
\begin{equation}
u_{\mathrm{old}}(x)=c_0+\sum_{r=1}^{m}a_r
\cos(\omega_r^\top x+b_r), \qquad m=\max(8,2d).
\end{equation}
We sample $\omega_r\sim\mathcal N(0,I_d/d)$,
$b_r\sim\operatorname{Uniform}(0,2\pi)$,
$a_r\sim\mathcal N(0,1/m)$, and $c_0\sim\mathcal N(0,0.25^2)$, then normalize
the values on the static collocation points. SIPMNN replaces this target with
the normalized current network output every 500 Adam steps.
This initialization is independent of the reference
eigenfunctions, so the coarse FDM shift remains the only external spectral
information supplied to the neural solver.
\subsection{Coarse and reference discretizations}
\label{app:fdm-sim}

\begin{table}[h]
\centering
\caption{Numerical calculations used for shifts, training, and evaluation.}
\label{tab:discretizations-sim}
\begin{tabular}{p{0.29\linewidth}p{0.47\linewidth}p{0.14\linewidth}}
\toprule
Component & Configuration & Purpose \\
\midrule
Coarse FDM & Tensor grid with $n=5$ interior points per coordinate and
$h=(b-a)/(n+1)$ on a nonperiodic interval $[a,b]$ & Fixed shifts \\
Neural collocation & 50,000 static Latin hypercube points sampled separately
from the FDM grid & Training \\
Anharmonic and Double Well references & Independent one-dimensional FDM with
800 interior points and a second-order central difference scheme. The
$d$-dimensional values are sums of one-dimensional eigenvalues & Evaluation \\
Coupled Oscillator and Quantum Harmonic references & Analytic eigenvalues &
Evaluation \\
Random Gaussian references & Independent Galerkin calculation in a sine basis
& Evaluation \\
\bottomrule
\end{tabular}
\end{table}

The reference calculations are used only for evaluation and do not enter the
coarse FDM calculation or neural training.
\subsection{Additional reproducibility details}
\label{app:repro-sim}

\begin{table}[h]
\centering
\caption{Implementation and training configuration.}
\label{tab:hyperparams-sim}
\begin{tabular}{p{0.30\linewidth}p{0.62\linewidth}}
\toprule
Setting & Value \\
\midrule
Coarse operator & Second-order central-difference stencil \\
Coarse eigensolver & LOBPCG \citep{knyazev2001lobpcg}, tolerance $10^{-7}$,
at most 500 iterations \\
Random seed & 0 for all random number generators \\
Random Gaussian potential &
$V(x)=\sum_{\ell=1}^{8}a_\ell
\exp(-\|x-c_\ell\|_2^2/0.15^2)$, with
$c_\ell\sim\operatorname{Uniform}([0,1]^d)$ and
$a_\ell\sim\operatorname{Uniform}([-5,5])$ \\
Boundary condition & Map coordinates to $\widetilde x\in[-1,1]^d$ and multiply
the raw output by $\prod_j(1-\widetilde x_j^2)$ \\
Neural network architecture & $[d,40,40,40,40,1]$ at $d=5$ and \\
& $[d,80,80,80,80,1]$ at $d=10$ \\
Activation & tanh \\
Collocation set & 50,000 static Latin hypercube points \\
Optimization & 50,000 Adam steps, learning rate $10^{-3}$,
$(\beta_1,\beta_2)=(0.9,0.999)$, and $\epsilon=10^{-8}$ \\
Modes & Two, updated within the same training loop \\
SIPMNN target update & Every 500 steps \\
Hardware & One shared NVIDIA A100 GPU with a 12 GB GPU-memory allocation \\
\bottomrule
\end{tabular}
\end{table}

\subsection{Adaptive baseline}
\label{app:adaptive-sim}

Both adaptive configurations initialize the effective shift to zero and update
it every 2,500 Adam steps according to
\begin{equation}
\sigma_i^{(m+1)}=\lambda_{i,\mathrm{best}}^{(m)}-\sigma_i^{(m)}-2,
\end{equation}
where $\lambda_{i,\mathrm{best}}^{(m)}$ is the Rayleigh quotient at the
checkpoint with the lowest total loss available at update $m$. Adaptive IPMNN has no
sigma penalty. For Adaptive SIPMNN, the penalty becomes active after the first
shift update and uses target $\tau_i^{(m)}=\sigma_i^{(m)}+2$ and scale
$\max(|\sigma_i^{(m)}|,|\tau_i^{(m)}|,1)$.
Unlike the fixed-FDM configurations, these adaptive baselines
receive no external spectral estimate and must infer the relevant
spectral region entirely from quantities produced during neural training.
\subsection{Bounded penalty and gradient}
\label{app:penalty-sim}

Let
\begin{equation}
p=w_\sigma\left(\frac{\lambda_{\mathrm{RQ}}-\tau_i}{s_i}\right)^2,
\qquad b=5\,\operatorname{stopgrad}(\mathcal L_{\mathrm{IPM}}).
\end{equation}
The implementation uses
\begin{equation}
\alpha=\min\left(1,
\frac{b}{\max(\operatorname{stopgrad}(p),\varepsilon)}\right),
\qquad \mathcal L_\sigma=\alpha p.
\end{equation}
The quantities used to compute $\alpha$ are detached. Therefore, the forward
value satisfies $0\leq\mathcal L_\sigma\leq
5\mathcal L_{\mathrm{IPM}}$, while backpropagation uses
\begin{equation}
\nabla_\theta\mathcal L_\sigma
=\alpha\nabla_\theta p.
\end{equation}
This is a detached gradient rescaling rather than the derivative of a hard
minimum. A hard minimum with detached $b$ would give zero penalty gradient when
$p>b$. The implemented form instead retains the direction toward the target
and reduces its scale by $\alpha$ when the cap is active. For fixed FDM,
$\tau_i=\widehat\lambda_i^{\mathrm{FDM}}$ and
$\sigma_i=\widehat\lambda_i^{\mathrm{FDM}}-1$. The bound limits the influence
of a poor coarse estimate without enforcing it as a hard target.
\subsection{Deflation normalization}
\label{app:deflation-sim}

Before the current estimate for mode $j$ is used in deflation, its values on
the static collocation points are normalized by their discrete Euclidean norm.
Equation~\ref{eq:deflation-sim} uses the same inner product, so
$\inner{\widetilde u_j}{\widetilde u_j}=1$. At each iteration, mode $i$ uses
the current estimates of modes $j<i$. Consequently, an inaccurate coarse shift
or lower-mode estimate may propagate to higher modes through deflation.

\end{document}